# Open-plan Glare Evaluator (OGE):

# A Demonstration of a New Glare Prediction Approach Using Machine Learning Algorithms

Ayman Wagdy [1 *], Veronica Garcia-Hansen [1], Mohammed Elhenawy [2], Gillian Isoardi [3], Robin Drogemuller [4], and Fatma Fathy [5]

1 Queensland University of Technology (QUT), Creative Industries Faculty, School of Design, Brisbane, Australia

2 Queensland University of Technology (QUT), Centre for Accident Research and Road Safety – Queensland (CARRS-Q), Brisbane, Australia

3 Light Naturally, Brisbane, Australia

4 Queensland University of Technology (QUT), Science and Engineering Faculty, School of Built Environment, Brisbane, Australia

5 Ain Shams University, Faculty of Engineering, Department of Architecture Engineering, Cairo, Egypt

* Corresponding author: Ayman Wagdy (a.wagdy@qut.edu.au)

**ABSTRACT** Predicting discomfort glare in open-plan offices is a challenging problem. Although glare research has existed for more than 50 years, all current glare metrics have accuracy limitations, especially in open-plan offices with low lighting levels. Thus, it is crucial to develop a new method to predict glare more accurately. This paper is the first to adopt Machine Learning (ML) approaches in the prediction of glare. This research aims to demonstrate the validity of this approach by comparing the accuracy of the new ML model for open-plan offices (OGE) to the accuracy of the existing glare metrics using local dataset. To utilize and test this approach, Post-Occupancy Evaluation (POE) and High Dynamic Range (HDR) images were collected from 80 occupants (n=80) in four different open-plan offices in Brisbane, Australia. Consequently, various multi-region luminance values, luminance, and glare indices were calculated and examined as input features to train ML models. The accuracy of the ML model was compared to the accuracy of 24 indices which were also evaluated using a Receiver Operating Characteristic (ROC) analysis to identify the best cutoff values (thresholds) for each index in open-plan configurations. Results showed that the ML approach could predict glare with an accuracy of 83.8% (0.80 true positive rate and 0.86 true negative rate), which outperformed the accuracy of the previously developed glare metrics. OGE is applicable for open-plan office situations with low vertical illuminance (200 to 600 lux). However, ML models can be trained with more substantial datasets to achieve global model.

## 1. INTRODUCTION

Discomfort glare is the primary source of visual discomfort in daylit open-plan offices, and preventing it requires an accurate prediction model to successfully utilize natural light in this type of space. Several studies have demonstrated the significance of visual discomfort in open-plan offices through Post-Occupancy Evaluation (POE), where 56% and 49% of the occupants reported discomfort from daylight in Konis [1] and in Hirning, Isoardi and Cowling [2] respectively. Various glare metrics have been developed to predict glare zones and distinguish the disturbing feeling by setting numerical thresholds. Glare predictive metrics use mathematical equations based on contrast effect, saturation effect, or absolute threshold, or equations derived from empirical equations [3]. The equations are mainly constructed from the luminance and solid angle of the glare source, the adaptive level, and the position index, which do not usually consider other factors as suggested by [4]. The majority of these metrics were developed and tested under certain conditions which makes it difficult to extrapolate results to other conditions [3]. Also, the inherent subjectivity in glare responses makes glare prediction even more complicated. Temporal variables can randomly affect occupants' glare sensation. Through a field experiment, it was found that occupants became more tolerant of glare from daylight as the day progressed [5, 6]. Several other studies have tested the limitations of glare indices in predicting visual discomfort [7-10]. Hence, modifications or new metrics are continuously evolving based on new experiments that assess users' visual comfort. Still, limitations exist in these experiments such as being conducted under particular luminous environments and controlled settings, e.g. fixed window sizes and view direction [3].

Glare metrics are usually derived from subjective ratings and luminance-based measurements. Several studies correlated glare metrics with quantitative illuminance-based metrics to avoid the computationally expensive luminance renderings [11-13]. They have considered illuminance-based metrics as the most weighted factor for predicting discomfort glare, for example, the simplified version of the DGP metric developed by Wienold [11]. Mardaljevic, Andersen, Roy and Christoffersen [12]examined the relation between Useful Daylight illuminance (UDI) and Daylight Glare Probability (DGP), aiming to find the potential of using UDI as a proxy of DGP. Karlsen, Heiselberg, Bryn and Johra [13] suggested the use of simple vertical illuminance at the eye and horizontal illuminance at desk level to indicate glare perception; however, this approach cannot be used in office spaces where the lighting levels are low or when a contrast-based glare accrues.

Recently, a study by Wienold, Iwata, Sarey Khanie, Erell, Kaftan, Rodriguez, Yamin Garreton, Tzempelikos, Konstantzos, Christoffersen, Kuhn, Pierson and Andersen [3] evaluated 22 glare metrics regarding their performance and robustness for daylit workspaces. Experimental datasets were collected in 6 different countries to offer more general results. According to their statistical results, DGP was found to be the most robust with the highest performance for predicting glare which is only valid for daylit dominated spaces with a 0.74 True Positive Rate (TPR) and a 0.72 True Negative Rate (TNR) over the combined dataset. The TPR refers to the prediction rate of disturbing glare situations, and TNR refers to the prediction rate of no-glare situations and theses rates are detailed in Rodriguez, Yamín Garretón and Pattini [14]. The equations of the highest six ranked metrics were found to be based on the saturation effect.

On the other hand, metrics mainly based on contrast effect or empirical equations had lower performance and robustness. However, the spaces under study by Wienold, Iwata, Sarey Khanie, Erell, Kaftan, Rodriguez, Yamin Garreton, Tzempelikos, Konstantzos, Christoffersen, Kuhn, Pierson and Andersen [3] were daylit cellular offices, unlike open-plan offices where dimmer lighting conditions prevailed. Hirning, Isoardi and Garcia-Hansen [15] investigated discomfort glare in open office spaces of 5 green buildings through 493 surveys. They compared the occupant responses with glare indices, and they found that the prediction of glare indices was statistically correlated to discomfort; however, all indices underestimated discomfort glare. Thus, they developed a new index UGP, which is derived from the UGR glare metric, to determine the likelihood of being glare disturbed in open office spaces. They achieved an overall accuracy of 69% for predicting discomfort glare (0.49 TPR and 0.78 TNR). Although UGP was specifically optimized for open-plan offices, it is still evident that a 51% false-negative prediction rate of discomfort glare indicates low prediction accuracy according to Wienold, Iwata, Sarey Khanie, Erell, Kaftan, Rodriguez, Yamin Garreton, Tzempelikos, Konstantzos, Christoffersen, Kuhn, Pierson and Andersen [3]. Therefore, this research aimed to develop a new approach for accurate glare prediction, which can be adapted to any space type of specific luminous condition like open-plan offices.

The existing research on discomfort glare focuses particularly on statistical methods to derive a threshold for glare indices to discriminate between discomfort and comfortable lighting situations, which may not be the ideal method for this task. Therefore, this paper sheds new light on the potential advantages of using Machine Learning models to predict discomfort glare with higher accuracy than the statistical methods. In general, Machine Learning (ML) techniques are divided into supervised and unsupervised learning. Unsupervised is usually used for solving clustering problems, which require just unlabeled datasets for the machine learning algorithms to search for the best way to group and interpret the data. Unlike unsupervised learning, supervised machine learning algorithms can be used to develop a predictive model based on a classification or regression technique which requires labelled data [16, 17]. In this paper, we used the supervised learning technique (i.e. classifiers) to classify HDR images captured during POE where the true labels/responses were collected from the occupants of the open-plan offices.

We present an innovative method that shows how to adopt Machine Learning (ML) to learn the relationship between different features and human visual discomfort in open-plan offices. Machine learning techniques are suitable models for the glare prediction problem for four reasons:

- First is the stochastic nature of the input-output data, where it is possible to find two different subjects having different discomfort glare levels in the same daylit space. In other words, the response corresponding to any input predictors is a distribution rather than a single point in the response space.
- Second, the problem is multivariate, and the relationships between variables are nonlinear.
- Third, there is no closed mathematical form (model) that can be used to explain the relationship between the input predictors (i.e. various multi-region luminance values, luminance, illuminance, and glare indices) and the discomfort glare.
- Fourth, the ability of machine learning to learn from noise meant that all data points were used without needing to exclude any data that might typically be excluded for reasons like veiling luminance on screen as a glare source, and differences in glare sensibility or susceptibility of the subjects. This simplifies the experimental analysis of all the data collected.

Hence, this study focuses on proposing a novel approach of developing glare prediction models based on extracting features from HDR images using several techniques, including various multi-region luminance values, luminance, illuminance, and glare indices which were tested as input features for machine learning algorithms.

This research aims to demonstrate an entirely new approach to determining glare metrics using Artificial Intelligence (AI) such as machine learning and deep learning techniques.

## 2. RESEARCH OBJECTIVES

- Propose a new approach for predicting glare using machine learning models, including the optimization of the input features for training ML algorithms.
- Demonstrate the validity of this new approach by calculating the accuracy of machine learning models and compare them to the new thresholds of the existing glare metrics derived from ROC analysis.
- Demonstrate an application for the new approach by developing a model for low light open-plan office buildings and compile it into an easy-to-use tool called the Open-plan Glare Evaluator (OGE).

## 3. METHODOLOGY

The methodology section describes the workflow used to develop new ways of developing predictive glare model for open-plan offices through machine learning as well as the current statistical methods usually used to derive thresholds for glare indices. First, the subjective user assessments were collected, which were coupled with HDR images taken at the head position during POE. This was followed by data processing to evaluate the HDR images in preparation for training the predictive models. As shown in Fig. 1, the methodology follows three parallel steps:

1) HDR images were tessellated into a smaller grid pattern, where the average luminance of each grid element was calculated and used as the input feature for ML training.

2) HDR images were evaluated using Evalglare to compute 24 luminance, illuminance, and glare scores that were then used as input features for ML training.

3) The 24 Evalglare outputs were used to conduct ROC curve analysis and to calculate a cut-off value for each glare metric specifically for open-plan offices.

First, the overall accuracy, as well as TPR and TNR in step 1 and 2 were compared and ranked to identify the best ML glare predictive model for open-plan offices. Then, in step 3, the same criteria were used to assess and rank the performance of 24 glare metrics using their calculated cut-off values. Finally, the selected ML glare predictive model was compared to the performance of Evalglare outputs (24 indices) in terms of overall accuracy, TPR, and TNR, as well as Area under the curve (AUC) and Squared Distance (SqD).

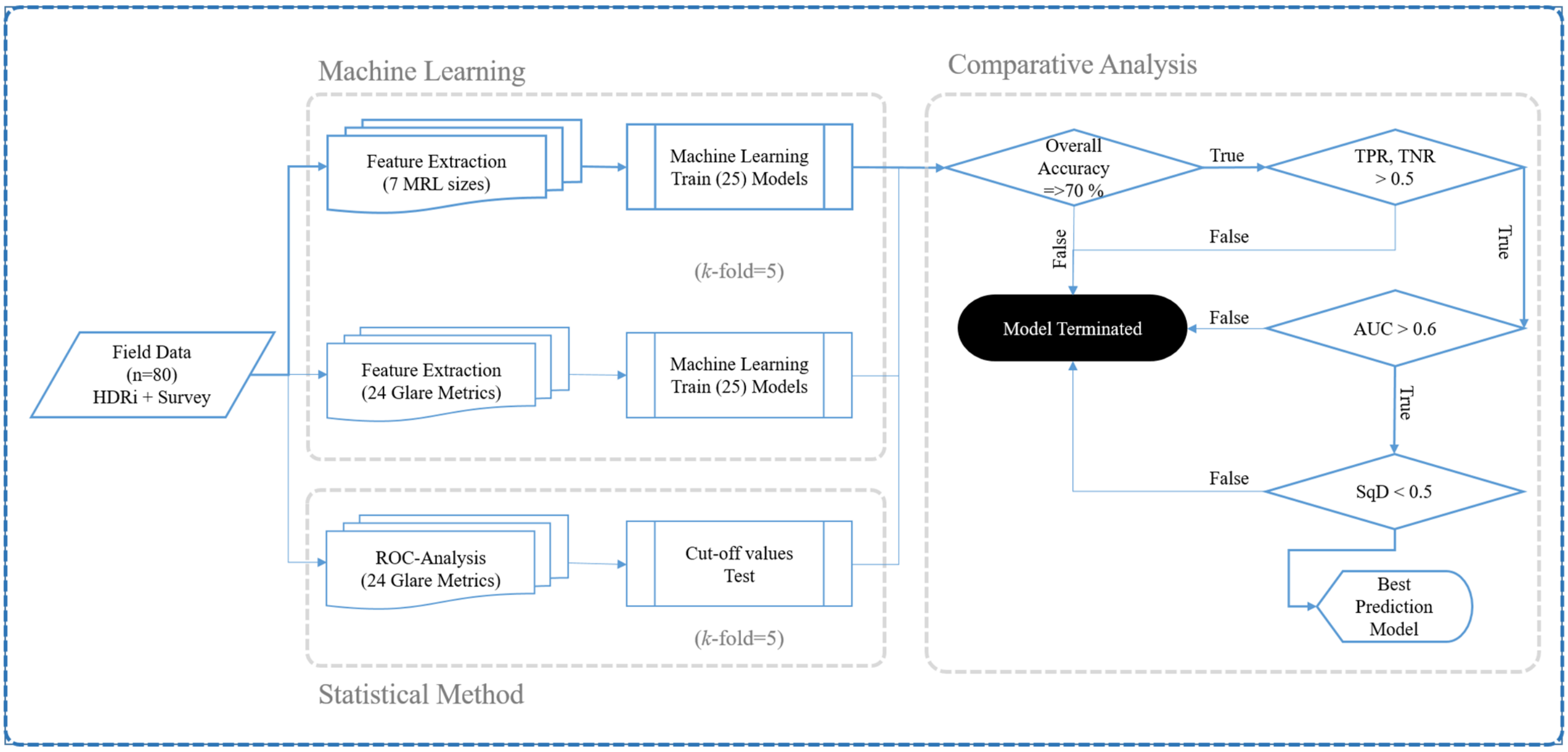

*FIGURE 1. Workflow of developing and analysing the predictive glare models.*

## 3.1. DATASET OVERVIEW: SURVEY + HDRI

Post occupancy evaluations (POE) were conducted on 80 subjects in four open-plan offices located in Brisbane, Australia. The users' assessments of their visual environments were collected through their responses based on the current lighting condition at their workstation. The number of occupants that experienced discomfort glare was 30 (37.5%), and it was 50 (62.5%) for those who reported no glare. At the same time, LDR images were captured to compute HDR images which were calibrated based on best practices [18-22]. The images were captured using a smartphone with a fisheye lens similar to the one presented in [18].

Vertical illuminance (Ev) was calculated from HDR images to show the variability in the lighting conditions. The vital statistical characteristics of Ev were represented in a violin plot (see Fig. 2) which confirms the low light levels usually found in open-plan offices since the majority of Ev values ranged from 200 to 600 lux.

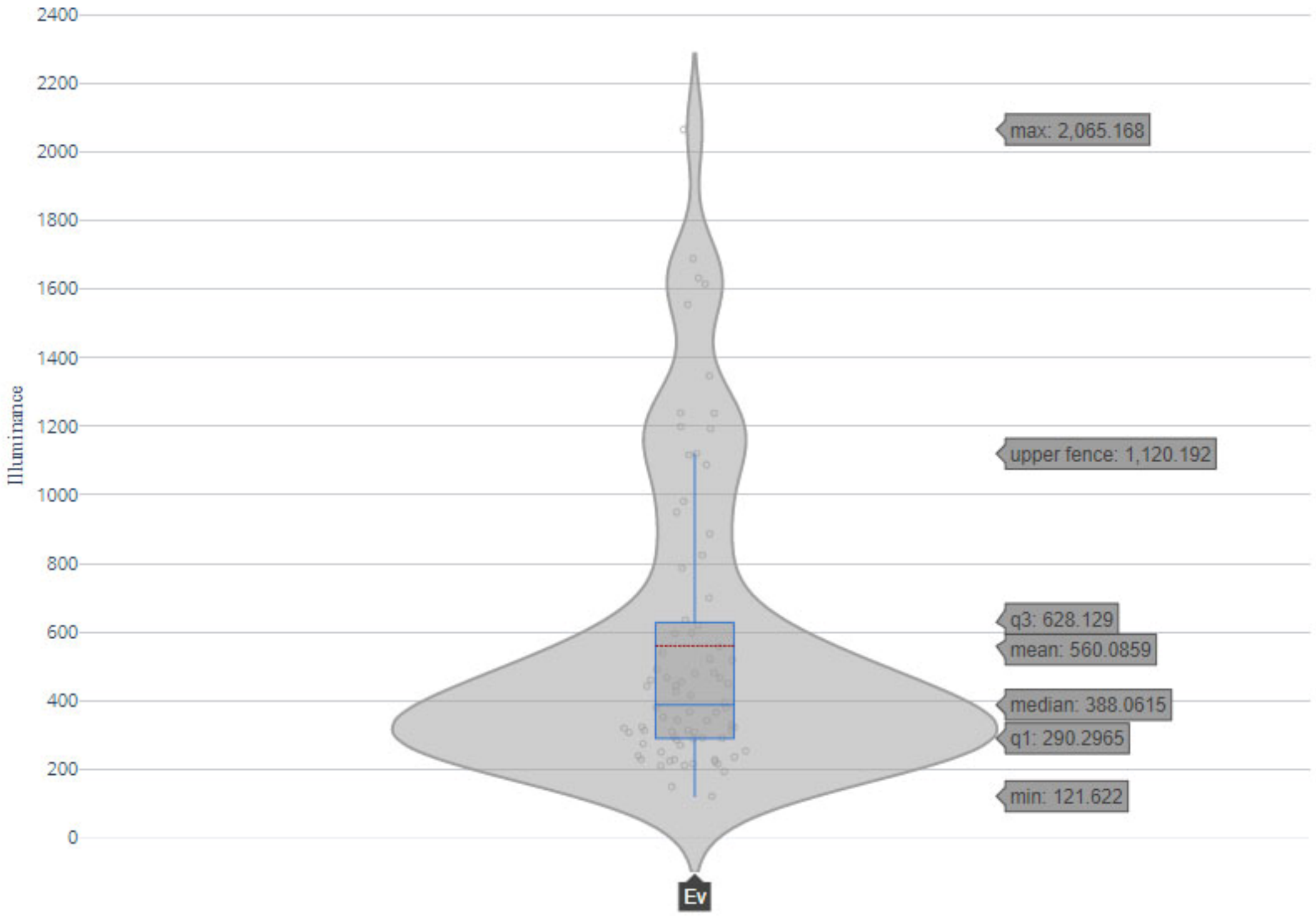

*FIGURE 2. The violin plot shows the distribution of vertical illuminance (Ev) values*

## 3.2. MACHINE LEARNING MODEL

Prior to training a machine learning model, the input data (features and responses) need to be acquired and formatted in a matrix D_(n *(m+1)), where n is the number of occupants (i.e. 80) and m is the number of the extracted features. First, features were extracted from HDR images and then coupled with the categorical responses of the occupant (Glare/No Glare), which was the last column in the matrix D. Two methods of feature extraction were explored in this research:

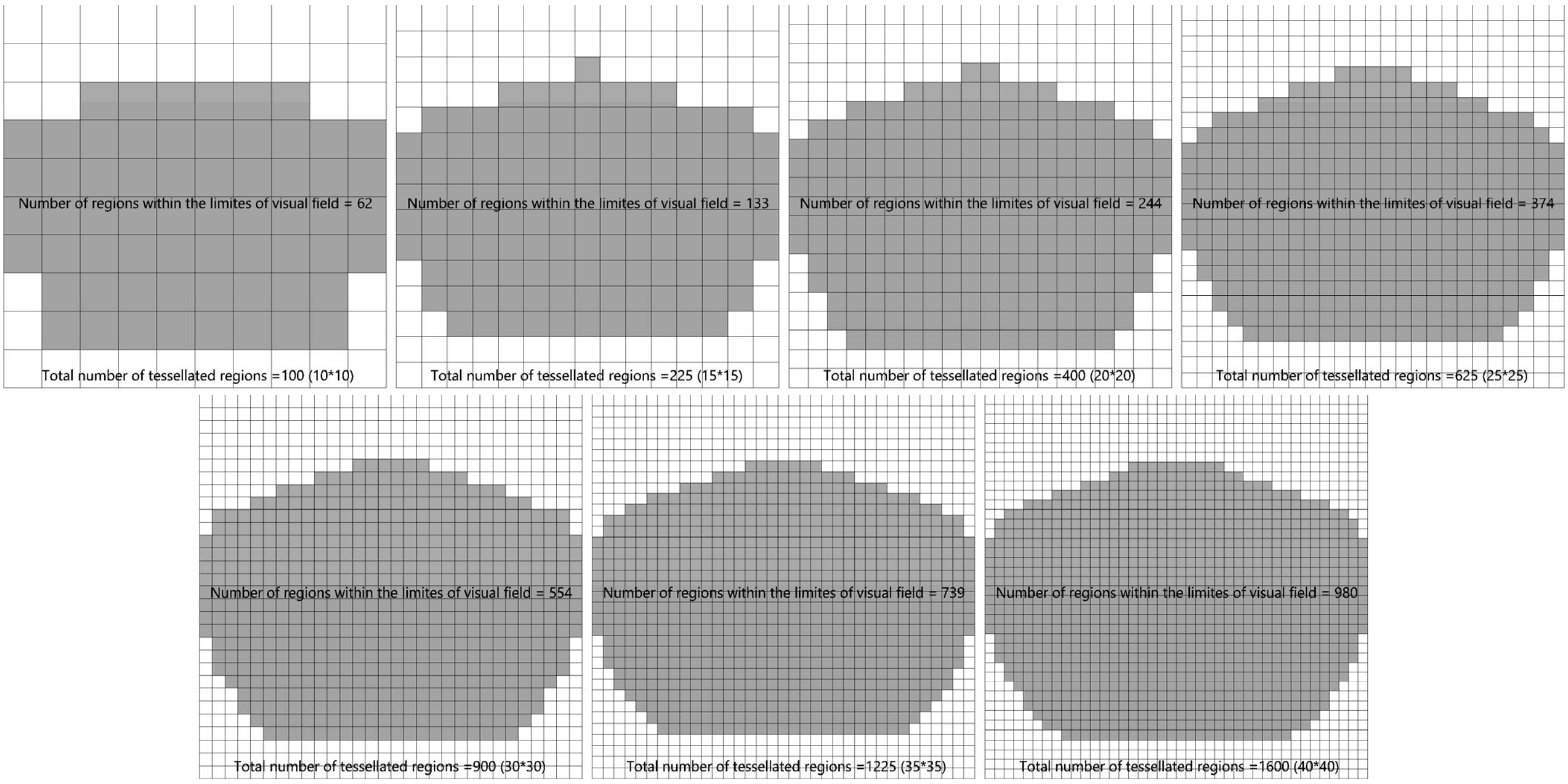

*FIGURE 3. The seven grid sizes of multi-region luminance highlighting the regions within the field of view*

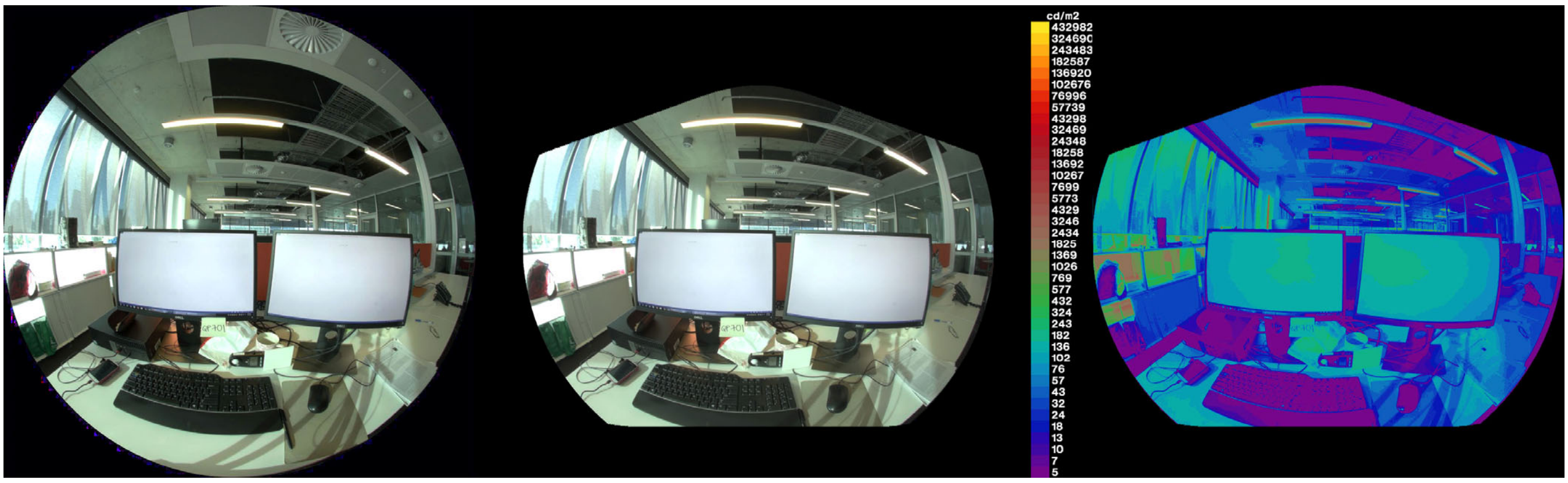

*FIGURE 4. The original view of the HDR image in the left, the cropped image according to human's field of view in the centre, and the false-colour map of the HDR image on the right*

### 3.2.1. Multi-Region Luminance (MRL)

The HDR images were tessellated into equally-sized regions at different resolutions to compute the average luminance for each region within the field of view. This method was previously proposed by Wagdy, Garcia-Hansen, Isoardi and Allan [23] as a multi-region contrast method to efficiently detect glare based on per-pixel luminance measurements of HDR images as it uses equidistance fisheye projection. In this study, seven (7) different sizes of the grid were parametrically generated using the brute-force method inside Grasshopper [24-26], starting from dividing the HDR images into 10 by 10 segments to 40 by 40 with an increment of five segments as shown in Table 1 and Fig. 3. Although averaging the luminance of the image reduces the details of the glare source, this aids the predictive algorithms as it minimises the overfitting problem. However, averaging the luminance with significant regions may result in losing the critical information needed by the prediction models to discriminate between glare and non-glare situations. Thus, it was essential to test different sizes to find a balance point between the region size and the prediction power.

The workflow of luminance extraction from the HDR images was repeated for each image, similar to Fig. 4. For a given grid, the average luminance of each region was calculated, as shown in Fig. 5. These averages and the corresponding occupant's response were concatenated to form a row vector in the matrix D. This process was repeated for all HDR images to construct the whole matrix D which was then named according to the reference code shown in Table 1. The seven Multi-Region Luminance (MRL) matrix files were analyzed one after another from MRL-62 (62 features) to MRL-980 (980 features).

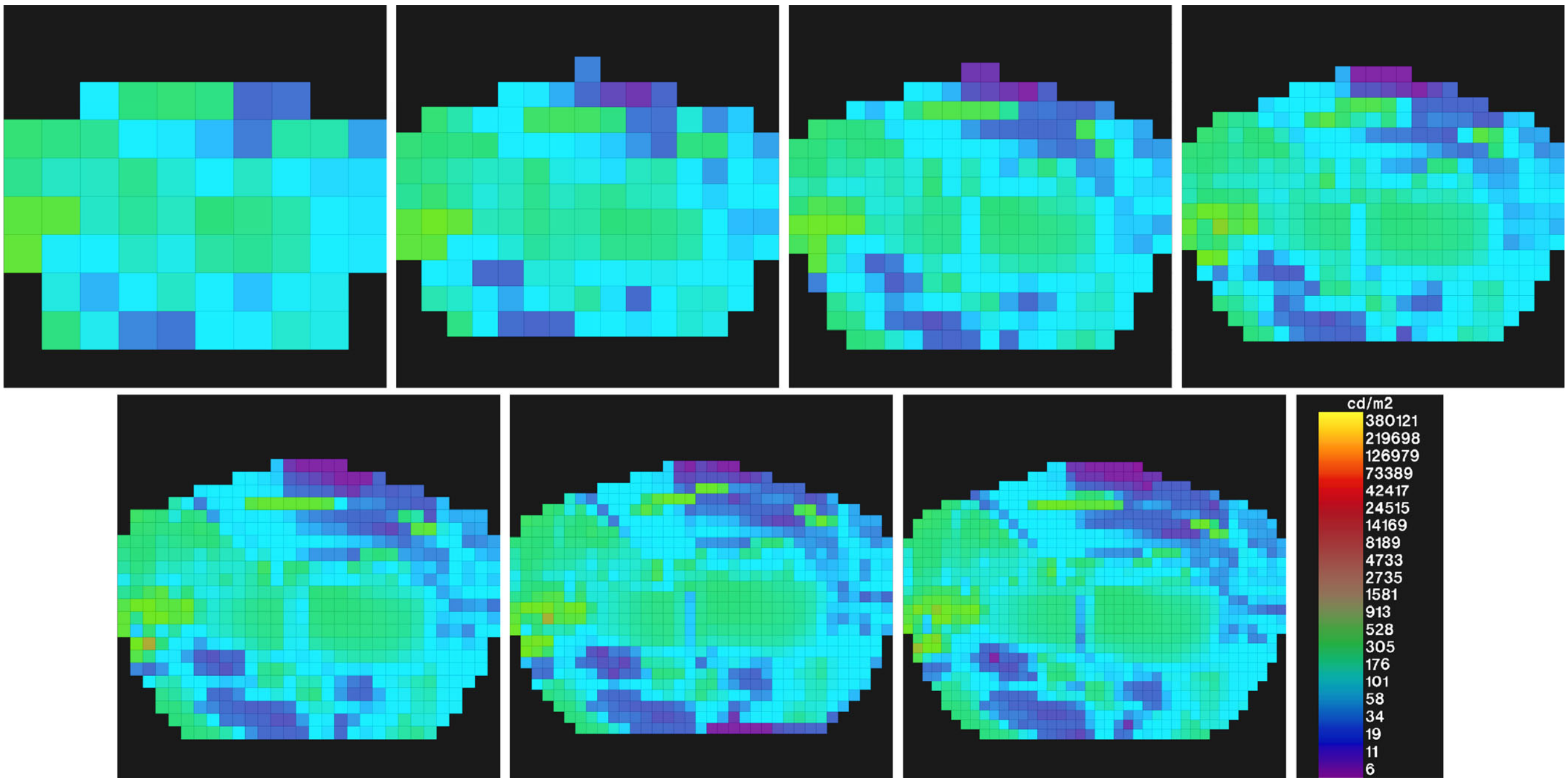

*FIGURE 5. Average luminance represented in each of the multi-region luminance grids.*

*TABLE 1. Description of the seven multi-region luminance grids*

| Grid size | Total number of regions | Number of regions within the field of view | Reference code of matrix ***D*** |
|---|---|---|---|
| **10*10** | 100 | 62 | MRL-62 |
| **15*15** | 225 | 133 | MRL-133 |
| **20*20** | 400 | 244 | MRL-244 |
| **25*25** | 625 | 374 | MRL-374 |
| **30*30** | 900 | 554 | MRL-544 |
| **35*35** | 1225 | 739 | MRL-739 |
| **40*40** | 1600 | 980 | MRL-980 |

### 3.2.2. Twenty-four (24) Glare Metrics

The second feature extraction method was done through Evalglare [27] which was used to conduct a detailed glare analysis in which 24 values of luminance, illuminance, and glare scores of multiple glare metrics, including the maximum and task luminance, were calculated and formatted in an 80 x 25 matrix (i.e. 24 features + response). The score of each of the 24 indices was then used as a predictive feature for the twenty-five ML classification models. Cut-off values of glare indices established in experimental scenarios may not be transferred practically to field settings [28], and previous research has indicated that one single metric may not be sufficient to predict glare experience with high accuracy [29]. Therefore, two sets of multiple glare indices were combined as groups of features to explore the possibility of getting better prediction accuracy when multiple metrics were used together. The first set included six metrics (DGP, DGI, UGR, VCP, CGI, and Lveil) which are the default outputs of Evalglare. The second set included the 22 glare, luminance, and illuminance outputs of Evalglare plus the task and maximum luminance when the detailed option in Evalglare was activated.

These results were combined with the subjective responses of the occupants (n=80) regarding their lighting conditions as a matrix of input features of HDR images. These data were imported into the classification learner app in MATLAB. A cross-validation scheme was applied to these data by randomly partitioning the data into five folds (k-fold = 5) to estimate the accuracy of the predictive model over each fold following the recommended settings by Rodriguez, Perez and Lozano [30]. In 5-fold cross-validation, the 80-row vectors were divided into five data blocks, each consisting of 16-row vectors. Then the ML model was trained using four data blocks and consequently was tested using the fifth remaining data block. This process was repeated five times, with each of the five data blocks used precisely once to test the model. The five testing results were then averaged to produce a single estimation for the model performance. Twenty-five (25) classification models in MATLAB were evaluated to identify the most accurate algorithm for classifying the (Glare/No-Glare) situation based on field data while using a different set of features each time. These algorithms included decision trees, discriminant analysis, support vector machines, logistic regression, nearest neighbors, naive bayes, and ensemble classification, which are explained in detail in [16, 17].

## 3.3. ROC CURVE ANALYSIS

After the training phase was carried out, all ML models, as well as the 24 luminance, illuminance, and glare metric scores, were used to generate Receiver Operating Characteristic (ROC) curves to evaluate the performance of each prediction model and to derive the optimal cut-off value for the 24 indices. The cut-off value for each metric was the value that discriminated between the glare and no-glare situation. Five-fold cross-validation was used in generating ROC to avoid any biasing that may accrue due to splitting the data randomly. The benefit of using this technique is to maximize the training and testing datasets, as recommended by Rodriguez, Perez and Lozano [30]. The five testing cut-off values of each metric were calculated and then averaged to produce a single estimation for the optimal cut-off value for each glare metric derived for open-plan offices. In addition to dividing the data (n=80) into five-folds, the data were treated as one combined dataset to get a broader range of lighting conditions. The cut-off values derived from the 5-fold trained (C1) and combined (C2) datasets were compared to find the cut-off variation between the two datasets and to indicate the robustness of the model. The smaller the difference, the more likely the same accuracy can be obtained when applying this cut-off for other datasets. The variation error (E) was calculated using equation 1.

$$E = \frac{|C1 - C2|}{C1} \times 100 \qquad (1)$$

Then, the prediction accuracy of each glare metric was compared to the accuracy of the best ML model in terms of overall accuracy, TPR, and TNR. The Overall Accuracy (OA) was calculated based on equation (2) where TP is the number of true positive cases indicating the number of correct predictions of glare situations, TN is the number of true negative cases indicating the number of correct predictions of no-glare situations, FP is the number of false-positive cases, and FN is the number of false-negative cases representing incorrect predictions of glare and no-glare situations respectively. TPR and TNR were also calculated to avoid misleading performance interpretations that may be caused by OA alone.

$$OA = \frac{(TP+TN)}{(TP+FP+TN+FN)} \qquad (2)$$

Further diagnostic tests [14] were conducted in which the Area Under the Curve (AUC), and the squared distance (SqD) were calculated. AUC is the area under the ROC curve, and it is used as a measure to indicate the ability of the model to discriminate glare situations. According to Hosmer Jr, Lemeshow and Sturdivant [31], if AUC equals or is larger than 0.8, it has an excellent discrimination ability, whereas 0.7 and 0.6 can be interpreted as having good and fair discrimination respectively [30]. The Squared Distance (SqD) is the length of the shortest line from the ROC curve to the top left corner point where TPR and TNR equal one. If SqD equals zero, it means that the model has 100% prediction accuracy. Therefore, the smaller the SqD, the higher the prediction accuracy it delivers. This value was computed for the trained 5-fold data as well as for the combined data.

### 3.4. PERFORMANCE ASSESSMENT CRITERIA

In order to assess the performance of the glare predictive models, all results were saved in multiple tables to evaluate the accuracy of each model relative to the feature used. Successful models were assumed to pass the following criteria:

- Overall prediction accuracy of 70% or higher. Since this paper focused on open-plan offices, our proposed model should achieve better performance than the current related glare metric, which had 69% overall accuracy Hirning, Isoardi and Garcia-Hansen [15].
- Both the True Positive Rate (TPR) and the True Negative Rate (TNR) need to be larger than 0.5 in order to be considered successful as a potential model for predicting glare. Thus, TPR and TNR lower than 0.5 was a terminating threshold to avoid random prediction results following the proposed performance criteria by Wienold, Iwata, Sarey Khanie, Erell, Kaftan, Rodriguez, Yamin Garreton, Tzempelikos, Konstantzos, Christoffersen, Kuhn, Pierson and Andersen [3].
- AUC equal to or larger than 0.6 was considered acceptable following Wienold, Iwata, Sarey Khanie, Erell, Kaftan, Rodriguez, Yamin Garreton, Tzempelikos, Konstantzos, Christoffersen, Kuhn, Pierson and Andersen [3] criteria.
- The SqD was considered acceptable with values smaller than 0.5.

Since this paper proposes new cut-off values for glare metrics which can increase the accuracy of predicting glare in open-plan offices, an additional variation error criterion was evaluated to the generality of the new cut-off values. Any model with a variation error below 10% indicated that this cut-off value could be generalized and extrapolated on other datasets for this type of office space.

# 4. RESULTS

## 4.1. MACHINE LEARNING MODEL

Results have shown a high potential for machine learning algorithms to predict glare across most of HDR images' features that represented multi-region luminance and glare scores. The overall accuracy matrix of 825 trained ML models was calculated. Each algorithm provided different accuracy for different features with an overall prediction accuracy ranging from 30% to 83.8%. Twelve (12) out of the 33 features failed to reach the assigned criterion which includes glare metrics, including DGP, UGR, and CGI which reached a maximum accuracy of 69%, 68%, and 65%, respectively. On the other hand, models that were trained with luminance-based indices like the Multi-Region Luminance, background luminance (lum background), average luminance (Av_Lum), median luminance of image (Med_lum), and size of the glare source (Omega_S) showed higher overall performance and more significant numbers of successful models using multiple ML algorithms.

The total number of ML models that succeeded in passing the first criterion across all features was 164 models. Some ML models performed better than the others; therefore, they were filtered according to the maximum OA acquired by each feature, as shown in Table 2, where TPR and TNR were then analyzed.

When combining the six default glare metrics as features in the same training model, no significant improvement was noticed, and only one model reached an overall accuracy of 70%. In the second combination of glare metrics, when all 24 glare indices were used, higher overall accuracies were achieved; the maximum (73.8%) was reached using Coarse Gaussian SVM [32]. However, by analyzing the TPR and TNR, it was found that TPR reached only 33% whereas TNR was 98% which indicated the underestimation of glare cases by 67%.

In multi-region luminance MRL, all grid sizes reached high accuracy using various ML algorithms. By analyzing TPR and TNR, most of them passed the assigned threshold. Only in MRL-62 and MRL 244, did the TPR fail to exceed 0.5, which indicated that glare was under-predicted. The highest accuracy of all models (83.8%) was found to be achieved by MRL-374 using RUS Boosted Trees algorithms [33]. In the case of MRL-374, both TPR and TNR values were higher than 50%; they reached 80 and 86%, respectively. The ROC curve was then generated to evaluate AUC and SqD. It was found that this model has an excellent discrimination ability since the calculated AUC was equal to 0.85 with a very small SqD of 0.06, as shown in Fig. 6. It outperformed the previously developed metrics for open-plan offices with higher accuracy of more than 14% and without any bias towards glare or no glare. Therefore, the trained ML model with RUS Boosted Trees algorithms using MRL-374 features was selected as the best ML glare predictive model for open-plan offices. This ML model was developed as a tool for glare assessment in open-plan offices with a friendly user interface which was named as the Open-plan Glare Evaluator (OGE) as shown in Fig. 7. In other terms, OGE tool does not require any knowledge of programming or machine learning to use. This tool takes the HDR images as input, extracts the luminance information of the 374 regions, and feeds them to the OGE model to predict the likelihood of glare.

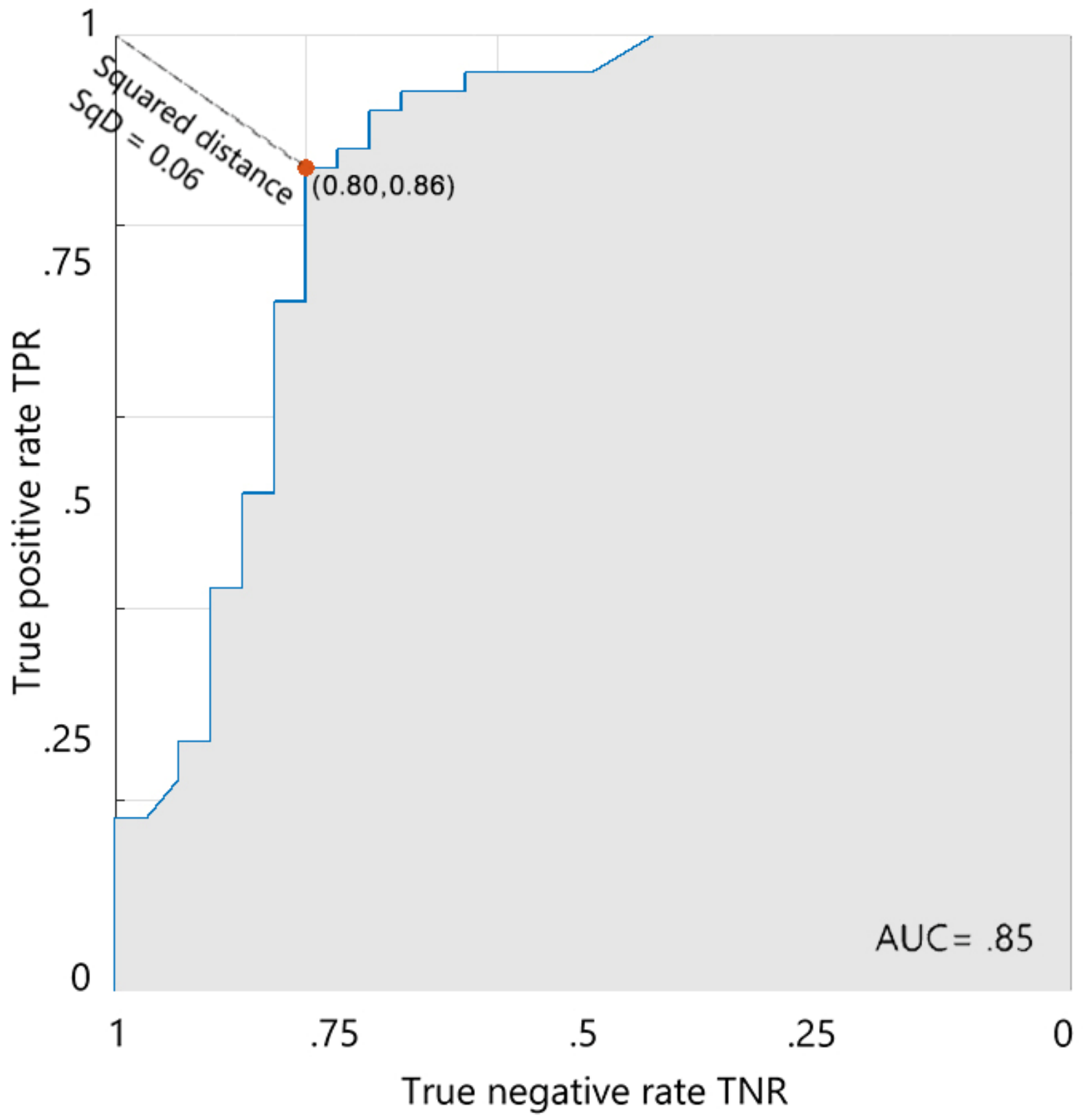

*FIGURE 6. Receiver Operating Characteristic (ROC) curve for the best performance of the ML models (RUS Boosted Tree algorithm with MRL-374 predictive features).*

*TABLE 2. The highest-ranked ml model based on the overall accuracy for each feature was extracted from table 1 and is detailed along with the corresponding TPR, TNR, and ml algorithm used. White marked cells show the models which have an overall accuracy of 70% or higher with TPR and TNR larger than or equal to 0.5 based on non–rounding values.*

| Features | Overall Accuracy | TPR | TNR | ML Algorithms |
|---|---|---|---|---|
| Ev | 70 | 0.33 | 0.92 | CoarseGaussianSVM |
| Ev_dir | 68.8 | 0.67 | 0.70 | SubspaceKNN |
| DGP | 68.8 | 0.47 | 0.82 | MediumGaussianSVM |
| UGP | 63.7 | 0.23 | 0.88 | FineGaussianSVM |
| UGR | 67.5 | 0.27 | 0.92 | FineGaussianSVM |
| UGR_exp | 65 | 0.3 | 0.88 | GaussianNaiveBayes |
| VCP | 65 | 0.37 | 0.82 | GaussianNaiveBayes |
| DGI | 70 | 0.4 | 0.88 | CoarseTree |
| DGI_mod | 70 | 0.4 | 0.88 | FineGaussianSVM |
| CGI | 65 | 0.4 | 0.80 | MediumKNN |
| DGR | 71.3 | 0.47 | 0.86 | KernelNaiveBayes |
| Lveil | 66.3 | 0.23 | 0.92 | MediumGaussianSVM |
| Lveil_CIE | 71.3 | 0.47 | 0.86 | KernelNaiveBayes |
| Omega_S | 71.3 | 0.37 | 0.92 | SubspaceDiscriminant |
| Lum_sources | 67.5 | 0.23 | 0.94 | MediumGaussianSVM |
| Av_Lum_pos | 68.8 | 0.33 | 0.90 | SubspaceDiscriminant |
| Av_Lum_pos2 | 68.8 | 0.5 | 0.80 | KernelNaiveBayes |
| Med_lum | 76.3 | 0.43 | 0.96 | SubspaceDiscriminant |
| Med_lum_pos | 73.8 | 0.43 | 0.92 | MediumKNN |
| Med_lum_pos2 | 77.5 | 0.43 | 0.98 | SubspaceDiscriminant |
| Av_Lum | 75 | 0.37 | 0.98 | LinearSVM |
| Lum Background | 76.3 | 0.47 | 0.94 | MediumGaussianSVM |
| Task Lum | 71.3 | 0.67 | 0.74 | FineTree |
| Max Lum | 65 | 0.37 | 0.82 | SubspaceDiscriminant |
| 6 Glare Metrics | 70 | 0.43 | 0.86 | LinearSVM |
| 24 Glare Metrics | 73.8 | 0.33 | 0.98 | CoarseGaussianSVM |
| MRL-62 | 77.5 | 0.5 | 0.94 | GaussianNaiveBayes |
| MRL-133 | 76.3 | 0.57 | 0.88 | GaussianNaiveBayes |
| MRL-244 | 72.5 | 0.47 | 0.88 | GaussianNaiveBayes |
| MRL-374 | 83.8 | 0.80 | 0.86 | RUSBoostedTrees |
| MRL-544 | 73.8 | 0.63 | 0.80 | RUSBoostedTrees |
| MRL-739 | 78.8 | 0.63 | 0.88 | BaggedTrees |
| MRL-980 | 76.3 | 0.53 | 0.90 | BaggedTrees |

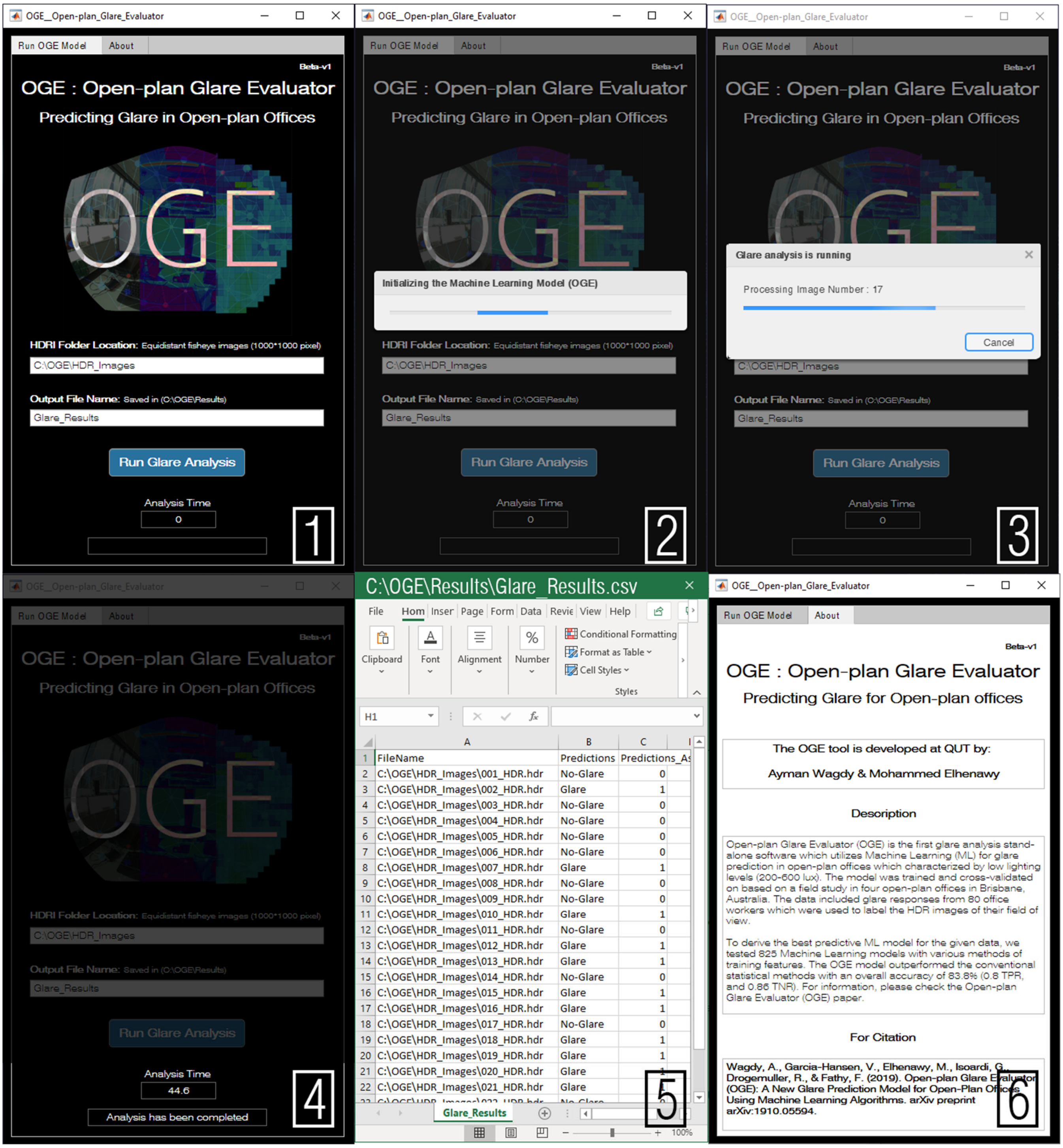

*FIGURE 7. The interface of OGE showing the process of predicting glare from HDR images to the output CSV file with glare prediction, design with an easy-to-use interface for architects and lighting researchers without a programming background.*

## 4.2. ROC RESULTS FOR GLARE METRICS

In order to compare the predictive power of the selected training ML model along with the existing glare metrics, all previously used glare metrics were evaluated using ROC analysis to evaluate their prediction accuracy using the same field data. A fair comparison can be achieved by randomly dividing the data into five folds (k-fold=5), similar to the data processing method used for ML. The accuracy and the overall cut-off value of each glare metric were obtained by averaging the accuracy results and the cut-off values from the five tests. Although the UGP metric was developed for open-plan offices, the overall accuracy was only 52% with TPR and TNR of 0.57 and 0.51, respectively, based on our dataset. The highest prediction accuracy was reached by the luminance background metric, where TPR and TNR were 0.66 and 0.89 with an overall accuracy of 78.75% and AUC and SqD were 0.68 and 0.22, respectively, as shown in Table 3 (left). When using the combined data, more glare metrics passed the assigned criteria, as shown in Table 4 where metrics were sorted based on the overall accuracy. In addition to luminance background, average luminance position, medium luminance position, and vertical illuminance showed potential for use in open-plan offices. By comparing the cut-off values for the five-fold dataset and the combined data, small variation errors were noticed in most of glare metrics which ranged from 0.1% in DGI and DGI modified to 7.3% in Luminance maximum. No variation was found between the two datasets in three metrics: Luminance background, Medium Luminance, and DGP. Thus, the assigned criteria (E<10%) was met by most of the glare metrics, except Lum_Sources, Lveil_cie, and Lveil.

## 5. DISCUSSION

Large open-plan offices, which are dim and mainly depend on electric lighting, have different lighting conditions than cellular offices and this is confirmed by the majority of Ev values (200 to 600 lux) collected during the field study. Thus, most of the available glare metrics studies and their cut-off values are not suitable for this type of office since most of these metrics are developed in experimental setups that simulate daylight-dominated cellular offices where occupants sit next to the window. Therefore, this study investigates three methods aiming to achieve the best predictive model for open-plan offices.

Based on the literature, the UGP metric, which is derived from UGR, is optimised explicitly for open-plan offices and reached an overall prediction accuracy of 69% (0.49 TPR and 0.78 TNR) according to Hirning, Isoardi and Garcia-Hansen [15]. However, 49% accuracy in detecting glare cases is not high enough to reduce random results. By applying the ROC curve analysis over the score of each index, new cut-off values are proposed. More insights are depicted by plotting the scores of each index and colour-coding them into red colour (Glare) and blue colour (No-Glare) based on the 80 survey responses, then deriving cut-off values from Table 3. It is evident in Fig. 8 that no cut-off value can totally discriminate glare situations which confirms the non-linear relationship between the glare score and people's perception of glare. However, a cut-off is derived in this paper which is optimized to reduce the number of incorrectly predicted cases by striking a balance between the true positive and true negative rates. By ranking the indices based on their performance, we notice that 4 out of the top 5 indices are luminance-based while the vertical illuminance Ev is ranked 4th. In term of glare metrics, DGI performs better than other glare metrics after UGR_exp.

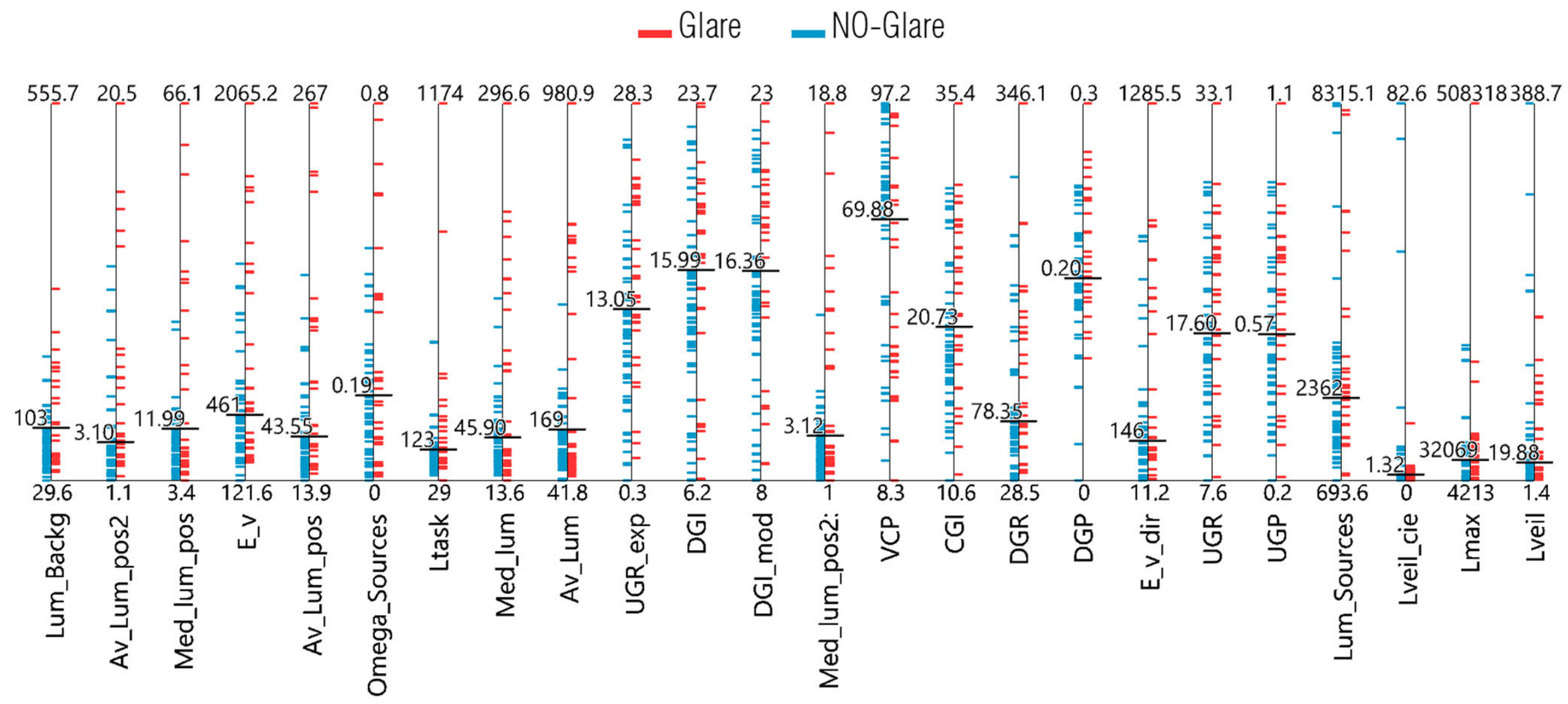

*FIGURE 8. The normalized score values for twenty-four indices with cut-off values.*

*TABLE 3. Performance of glare metrics using statistical method through averaging fivefold data (left) and using all data combined (right). Cut-off values of the 24 metrics derived from the average of the testing data on the left and compared to the cut-off values calculated from all data combined on the right. The variation error is calculated, which proves that the proposed cut-off value for each glare metric derived from testing data can be used to discriminate between glare and no-glare situations. White marked cells show the models which have an overall accuracy of 70% or higher with TPR and TNR larger than or equal to 0.5 based on non–rounding values.*

| | Average of Testing Data (k-fold=5) | | | | | | All Data Combined | | | | | | |
|---|---|---|---|---|---|---|---|---|---|---|---|---|---|
| **Glare Metric** | **Overall Accuracy** | **TPR** | **TNR** | **Cut-off** | **AUC** | **SqD** | **Overall Accuracy** | **TPR** | **TNR** | **Cut-off** | **AUC** | **SqD** | **%Error** |
| **Lum_Backg** | **78.75** | **0.66** | **0.89** | **103** | **0.68** | **0.22** | 76.25 | 0.57 | 0.88 | 103 | 0.69 | 0.20 | 0 |
| Av_Lum_pos2 | 62.5 | 0.65 | 0.64 | 3.10 | 0.70 | 0.20 | 71.25 | 0.60 | 0.78 | 3.34 | 0.69 | 0.21 | 7.7 |
| Med_lum_pos | 68.75 | 0.51 | 0.83 | 11.99 | 0.65 | 0.27 | 71.25 | 0.53 | 0.82 | 12.11 | 0.66 | 0.25 | 1 |
| E_v | 66.25 | 0.63 | 0.70 | 461 | 0.68 | 0.19 | 70 | 0.63 | 0.74 | 466 | 0.68 | 0.20 | 1.1 |
| Av_Lum_pos | 63.75 | 0.58 | 0.70 | 43.55 | 0.69 | 0.22 | 70 | 0.60 | 0.76 | 42.72 | 0.68 | 0.22 | 1.9 |
| Omega_Sources | 61.25 | 0.57 | 0.66 | 0.19 | 0.65 | 0.24 | 70 | 0.57 | 0.78 | 0.20 | 0.63 | 0.23 | 5.3 |
| Task Lum | 61.25 | 0.49 | 0.71 | 123 | 0.72 | 0.21 | 68.75 | 0.60 | 0.74 | 116 | 0.69 | 0.23 | 5.7 |
| Med_lum | 67.5 | 0.69 | 0.68 | 45.90 | 0.65 | 0.25 | 67.5 | 0.63 | 0.70 | 45.88 | 0.67 | 0.23 | 0 |
| Av_Lum | 61.25 | 0.50 | 0.70 | 169 | 0.65 | 0.24 | 67.5 | 0.57 | 0.74 | 161 | 0.66 | 0.25 | 4.7 |
| UGR_exp | 52.5 | 0.54 | 0.54 | 13.05 | 0.65 | 0.26 | 67.5 | 0.57 | 0.74 | 13.79 | 0.63 | 0.25 | 5.7 |
| DGI | 56.25 | 0.54 | 0.60 | 15.99 | 0.61 | 0.22 | 66.25 | 0.60 | 0.70 | 15.97 | 0.59 | 0.25 | 0.1 |
| DGI_mod | 58.75 | 0.54 | 0.64 | 16.36 | 0.62 | 0.22 | 66.25 | 0.60 | 0.70 | 16.38 | 0.6 | 0.25 | 0.1 |
| Med_lum_pos2 | 66.25 | 0.59 | 0.72 | 3.12 | 0.67 | 0.26 | 66.25 | 0.57 | 0.72 | 3.04 | 0.68 | 0.26 | 2.6 |
| VCP | 56.25 | 0.57 | 0.57 | 69.88 | 0.65 | 0.24 | 65 | 0.63 | 0.66 | 69.09 | 0.63 | 0.25 | 1.1 |
| CGI | 57.5 | 0.57 | 0.60 | 20.73 | 0.64 | 0.25 | 65 | 0.60 | 0.68 | 20.99 | 0.62 | 0.26 | 1.3 |
| DGR | 55 | 0.57 | 0.55 | 78.35 | 0.65 | 0.24 | 65 | 0.60 | 0.68 | 84.77 | 0.62 | 0.26 | 8.2 |
| DGP | 55 | 0.59 | 0.53 | 0.20 | 0.67 | 0.23 | 62.5 | 0.70 | 0.58 | 0.20 | 0.67 | 0.27 | 0 |
| E_v_dir | 53.75 | 0.54 | 0.55 | 146 | 0.63 | 0.26 | 62.5 | 0.63 | 0.62 | 140 | 0.63 | 0.28 | 4.1 |
| UGR | 52.5 | 0.57 | 0.51 | 17.60 | 0.63 | 0.25 | 61.25 | 0.67 | 0.58 | 17.32 | 0.61 | 0.29 | 1.6 |
| UGP | 52.5 | 0.57 | 0.51 | 0.57 | 0.63 | 0.25 | 61.25 | 0.67 | 0.58 | 0.56 | 0.61 | 0.29 | 1.8 |
| Lum_Sources | 52.5 | 0.59 | 0.49 | 2362 | 0.57 | 0.34 | 61.25 | 0.50 | 0.68 | 2623 | 0.58 | 0.35 | 11 |
| Lveil_cie | 53.75 | 0.44 | 0.62 | 1.32 | 0.57 | 0.33 | 60 | 0.53 | 0.64 | 1.08 | 0.55 | 0.35 | 18.2 |
| Max Lum | 57.5 | 0.71 | 0.49 | 32069 | 0.66 | 0.31 | 58.75 | 0.67 | 0.54 | 29737 | 0.66 | 0.32 | 7.3 |
| Lveil | 56.25 | 0.54 | 0.60 | 19.88 | 0.62 | 0.28 | 58.75 | 0.60 | 0.58 | 16.14 | 0.61 | 0.33 | 18.8 |

To validate the use of ML models in glare evaluation, the prediction accuracy of each model is compared to the accuracy derived from conventional statistical methods. Therefore, ROC analysis is applied to obtain cut-off values, AUC, and SqD of all glare metrics that were used previously as features in ML models. Background luminance achieves the highest prediction accuracy (78.75%) among all glare metrics computed by Evalglare detailed analysis. The same field data of ML models are used for ROC analysis in two ways; in the first, the average performance is computed from 5 data folds to resemble the k-fold method of ML model training and thus can ensure a fair comparison between both methods. In the second, all data combined is used, and analysis results are compared with 5-fold by calculating percentage error. This provides an insight into the robustness of the proposed cut-off values for open-plan offices.

By comparing the DGP optimal cut-off values for open-plan offices (0.2) to the original threshold (0.4), which divides between noticeable and disturbing glare, it is found that the original cut-off value of underestimated glare of the dim daylighting environment confirms the findings by Wienold, Iwata, Sarey Khanie, Erell, Kaftan, Rodriguez, Yamin Garreton, Tzempelikos, Konstantzos, Christoffersen, Kuhn, Pierson and Andersen [3] and Mahić, Galicinao and Van Den Wymelenberg [34]. Having a 50% difference between the original threshold developed for this metric does not indicate the non-robustness of the metric as it is developed for cellular offices which have different light characteristics. UGP is the only metric specifically developed for open-plan offices; however, it only reaches 52% overall accuracy when using trained data and 61.25% using all data.

Vertical illuminance (Ev) reaches an overall accuracy of 66.25% (TPR 0.63 and TNR 0.70) when using the average of testing data (k-fold method). However, it passed the assigned criteria when testing all data combined. According to Wienold, Iwata, Sarey Khanie, Erell, Kaftan, Rodriguez, Yamin Garreton, Tzempelikos, Konstantzos, Christoffersen, Kuhn, Pierson and Andersen [3], an Ev cut-off value around 3300 lux is ranked 2nd for side-lit offices, and based on our study, an Ev cut-off value of 461 lux is ranked 4th. This shows that Ev can be related to the adaptation level of both studies. Ev is useful as it can be measured easily with an illuminance meter in a field or experimental setup or calculated through HDR images or Radiance simulations which are less computationally expensive luminance renderings [12-14]. Following the proposed method by Karlsen, Heiselberg, Bryn and Johra [13], Ev with a cut-off value of (461 lux) can be used as a fast glare detection threshold in open-plan offices. However, it needs further investigation to ensure its validity. Adapting machine learning ML indicates a strong potential for models with high overall accuracy since it can identify the non-linear relationships between developed luminance or glare indices and survey responses. Multiple ML algorithms succeed in predicting discomfort glare with high accuracy between 70% and 83.8%. This includes a wide range of well-known glare metrics and luminance scores used to train ML models after correlating their values with the corresponding survey responses collected from 80 subjects. Their actual thresholds were not used. Instead, they are used as features to train the ML algorithms to find its own pattern to classify the input data into glare/no-glare situations.

Table 4 shows a comparative analysis between machine learning and statistical approaches when using the 24 glare scores for glare prediction. It can be noticed that in machine learning, most glare indices achieve high overall accuracy; however, it is achieved by high TNR, whereas TPR is less than 0.5 except in task luminance. On the other hand, when applying the cut-off values, both TPR and TNR are unbiased, though the overall accuracy is not high enough. Nevertheless, using luminance background with a cut-off value of 103 can balance between required criteria.

Multi-Region Luminance MRL is used as a predictive feature in ML models and has succeeded in finding the non-linear relationship between glare responses and extracted features. Consideration should be taken on the applied resolution (number of regions) which represents the number of features. Accordingly, the smallest number of regions that obtain the highest accuracy is sought to avoid too detailed data that can cause overfitting. MRL of grid size 25 by 25 reaches the highest accuracy; however, further optimization could be beneficial to find more optimal results in sizes ranging between 20 and 30 with 1-step increments. In general, the MRL method achieves high prediction accuracy as it extracts contrast from luminance measurements which is a severe cause of glare in open-plan offices.

The best performance of the model is obtained when using Multi-Region Luminance MRL-374 as input features. This differs from other features in correlating multi luminance measurements rather than a single measurement of the glare indices.

The MRL method uses location characteristics similar to the position index. Therefore, in this study, various sizes are tested, ranging from 10 by10 (MRL-62) to 40 by 40 (MRL-980) with five increments, finding that 25 by 25 (MRL-374) delivers the highest performance using the RUS Boosted Trees algorithm [35]. Small numbers of regions in MRL-62 means averaging a large number of pixels to calculate the average luminance of the region. Thus, significant contrast ratios can be missed. By increasing the number of regions (small grid size), this issue can be resolved, which clarifies the high accuracy of MRL-374 (83.8%). However, by increasing the tessellation to a higher resolution with a more significant number of regions, the prediction accuracy decreases as in MRL-980 (76.3%). This is due to the incapability of the ML model to find the relationship of regions luminance values and glare response with a large number of regions and limited data (n=80).

Moreover, if the full image is used, overfitting problems will accrue as the ML will not be able to learn from 1 million parameters as the total number of labelled data is just 80. A new dimension reduction method may tackle this problem, such as the ones presented in Elhenawy, Masoud, Glaser and Rakotonirainy [36]. Finally, using a non-uniform grid may give a better description of small glare sources or reflections causing glare. However, this approach requires further investigation as the total number of regions will vary from one image to another, which will bring rise to another set of data management problems.

Although the overall accuracy of the trained models achieves high values, TPR and TNR are evaluated as well to detect if there is any glare bias. For example, in the case where all glare metrics are used as features, accuracy reaches 73.8%; however, TPR goes below 50% and is compensated by a high TNR (98%) which means it is biased toward no-glare situations. Another example is when using Ev as a predictive feature. It achieves better performance in ML; however, it comes with a high bias toward no-glare prediction (TNR=0.92) and misclassifies glare situations by 67% (TPR=0.33) as shown in Table 4.

*TABLE 4. Comparison between using the glare scores as features for ML models and using cut-off value based on ROC analysis*

| **Features (glare indices)** | **Machine Learning** | | | **ROC Analysis** | | | |
|---|---|---|---|---|---|---|---|
| | **Overall Accuracy** | **TPR** | **TNR** | **Overall Accuracy** | **TPR** | **TNR** | **Cut-off** |
| Ev | 70 | 0.33 | 0.92 | 66.25 | 0.63 | 0.70 | 461 |
| Ev_dir | 68.8 | 0.67 | 0.70 | 53.75 | 0.54 | 0.55 | 146 |
| DGP | 68.8 | 0.47 | 0.82 | 55 | 0.59 | 0.53 | 0.20 |
| UGP | 63.7 | 0.23 | 0.88 | 52.5 | 0.57 | 0.51 | 0.57 |
| UGR | 67.5 | 0.27 | 0.92 | 52.5 | 0.57 | 0.51 | 17.60 |
| UGR_exp | 65 | 0.3 | 0.88 | 52.5 | 0.54 | 0.54 | 13.05 |
| VCP | 65 | 0.37 | 0.82 | 56.25 | 0.57 | 0.57 | 69.88 |
| DGI | 70 | 0.4 | 0.88 | 58.75 | 0.54 | 0.64 | 16.36 |
| DGI_mod | 70 | 0.4 | 0.88 | 56.25 | 0.54 | 0.60 | 15.99 |
| CGI | 65 | 0.4 | 0.80 | 57.5 | 0.57 | 0.60 | 20.73 |
| DGR | 71.3 | 0.47 | 0.86 | 55 | 0.57 | 0.55 | 78.35 |
| Lveil | 66.3 | 0.23 | 0.92 | 56.25 | 0.54 | 0.60 | 19.88 |
| Lveil_CIE | 71.3 | 0.47 | 0.86 | 53.75 | 0.44 | 0.62 | 1.32 |
| Omega_S | 71.3 | 0.37 | 0.92 | 61.25 | 0.57 | 0.66 | 0.19 |
| Lum_sources | 67.5 | 0.23 | 0.94 | 52.5 | 0.59 | 0.49 | 2362 |
| Av_Lum_pos | 68.8 | 0.33 | 0.90 | 63.75 | 0.58 | 0.70 | 43.55 |
| Av_Lum_pos2 | 68.8 | 0.5 | 0.80 | 62.5 | 0.65 | 0.64 | 3.10 |
| Med_lum | 76.3 | 0.43 | 0.96 | 67.5 | 0.69 | 0.68 | 45.90 |
| Med_lum_pos | 73.8 | 0.43 | 0.92 | 68.75 | 0.51 | 0.83 | 11.99 |
| Med_lum_pos2 | 77.5 | 0.43 | 0.98 | 66.25 | 0.59 | 0.72 | 3.12 |
| Av_Lum | 75 | 0.37 | 0.98 | 61.25 | 0.50 | 0.70 | 169 |
| **Lum Background** | 76.3 | 0.47 | 0.94 | **78.75** | **0.66** | **0.89** | **103** |
| **Task Lum** | **71.3** | **0.67** | **0.74** | 61.25 | 0.49 | 0.71 | 123 |
| Max Lum | 65 | 0.37 | 0.82 | 57.5 | 0.71 | 0.49 | 32069 |

## 6. CONCLUSION

The novelty of using machine learning in glare prediction applications is demonstrated here to yield better prediction power for the data collected in this scenario than any previous experimental method, including all glare indices. This shows the potential for incorporating machine learning algorithms in glare analysis to help design more visually comfortable buildings. Our local machine learning model achieves higher prediction power (83.8% overall accuracy, with TPR of 0.8 and TNR of 0.86) than the previously developed glare metric for open-plan offices such as UGP which had only 69% overall accuracy (0.49 TPR and 0.78 TNR). This method showed a high potential for extending the use of ML models, after adding more field data of different light scenarios, to accumulate its learning patterns in order to make global models.

The proposed ML model, as well as the reported cut-off values of other glare metrics, are only valid to open-plan offices situations with overall low light levels in general. However, the same machine learning approach could also be applied to other space types with different luminous conditions like cellular offices which are typically daylight-dominated, and this will be investigated in a future study. In this paper, 80 survey responses were included in training the ML model. More field data that have already been taken in other glare studies can be incorporated in our developed model to evaluate the robustness of our predictive model and validate it on more space prototypes. It's evident that using this method as a global threshold to be widely applicable would require more data sets to be collected. However, there is value still in using small data sets to determine local solutions besides demonstrating the applicability of our proposed approach.

All glare metrics except task luminance are found not to be useful predictive features for training machine learning models even when combining different data sets. On the contrary, the adaptation of Multi-Region Luminance MRL as a predictive feature succeeded in explaining the relationship between the input predictors and the occupant's visual discomfort in open-plan offices.

It's evident that a model with 100% accuracy is not possible due to the fact that discomfort glare is a subjective phenomenon, and as such, it is challenging to assess through subjective qualitative methods accurately. However, we found that machine learning algorithms can deal with the stochastic nature of this subjective aspect in glare prediction with higher accuracy.

A crucial limitation of using current machine learning models is that they depend on selected features that are supplied to the models. Therefore, improper selections may negatively affect the prediction power of the model, which is experienced when using some glare scores as features. On the contrary, Deep Learning (DL) can learn directly from data (end-to-end learning workflow) which may achieve higher accuracy. However, developing a DL prediction model would require tens to hundreds of thousands of images to train such models, and this will be investigated in a future study.

The pre-trained ML model is compiled in the form of stand-alone software which not only allows the users to use the trained model to predict glare but also to train the model and develop new models using new HDR images. This software, Open-plan Glare Evaluator (OGE), as well as the MATLAB code, are freely available through GitHub and can be accessed through the following web link: https://github.com/aymanwagdy/OGE-OpenPlan_Glare_Evaluator.

## 7. Acknowledgments

The Australian Government funded this work through the Australian Research Council (ARC) linkage project awarded in 2015 [Project number LP150100179], in partnership with AECOM Australia and Light Naturally. The ARC project is about "Designing healthy and efficient luminous environments in Green Buildings". The research is supported as well by the faculty of Creative Industries, Queensland University of Technology (QUT) through PhD and top-up excellence scholarships awarded to the first author.

## 8. REFERENCES

[1] K. Konis, Evaluating daylighting effectiveness and occupant visual comfort in a side-lit open-plan office building in San Francisco, California, Building and Environment, 59 (2013) 662-677.
[2] M.B. Hirning, G.L. Isoardi, I. Cowling, Discomfort glare in open plan green buildings, Energy and Buildings, 70 (2014) 427-440.
[3] J. Wienold, T. Iwata, M. Sarey Khanie, E. Erell, E. Kaftan, R.G. Rodriguez, J.A. Yamin Garreton, T. Tzempelikos, I. Konstantzos, J. Christoffersen, T.E. Kuhn, C. Pierson, M. Andersen, Cross-validation and robustness of daylight glare metrics, Lighting Research & Technology, (2019) 1477153519826003.
[4] C. Pierson, J. Wienold, M. Bodart, Discomfort glare perception in daylighting: influencing factors, Energy Procedia, 122 (2017) 331-336.
[5] M.G. Kent, S. Altomonte, R. Wilson, P.R. Tregenza, Temporal effects on glare response from daylight, Building and Environment, 113 (2017) 49-64.
[6] M. Kent, S. Altomonte, P. Tregenza, R. Wilson, Discomfort glare and time of day, Lighting Research and Technology, 47 (6) (2015) 641-657.
[7] J.Y. Suk, M. Schiler, K. Kensek, Investigation of existing discomfort glare indices using human subject study data, Building and Environment, (2017).
[8] K. Van Den Wymelenberg, M. Inanici, A Critical Investigation of Common Lighting Design Metrics for Predicting Human Visual Comfort in Offices with Daylight, LEUKOS, 10 (3) (2014) 145-164.

[9] M.B. Hirning, G.L. Isoardi, S. Coyne, V.R. Garcia Hansen, I. Cowling, Post occupancy evaluations relating to discomfort glare: A study of green buildings in Brisbane, Building and Environment, 59 (2013) 349-357.
[10] V. Garcia-Hansen, A.C. Allan, G. Isoardi, A. Wagdy, S.S. Smith, Evaluating visual comfort in open-plan offices: Exploration of simple methods for evaluation and prediction, in: CIE 2017- Smarter Lighting for Better Life, Jeju, South Korea, 2017.
[11] J. Wienold, Dynamic simulation of blind control strategies for visual comfort and energy balance analysis, in: Building Simulation, 2007, pp. 1197-1204.
[12] J. Mardaljevic, M. Andersen, N. Roy, J. Christoffersen, Daylighting metrics: is there a relation between useful daylight illuminance and daylight glare probabilty?, in: Proceedings of the building simulation and optimization conference BSO12, 2012.
[13] L. Karlsen, P. Heiselberg, I. Bryn, H. Johra, Verification of simple illuminance based measures for indication of discomfort glare from windows, Building and Environment, 92 (2015) 615-626.
[14] R.G. Rodriguez, J.A. Yamín Garretón, A.E. Pattini, An epidemiological approach to daylight discomfort glare, Building and Environment, (2017).
[15] M.B. Hirning, G.L. Isoardi, V.R. Garcia-Hansen, Prediction of discomfort glare from windows under tropical skies, Building and Environment, 113 (2017) 107-120.
[16] T. Hastie, R. Tibshirani, J. Friedman, J. Franklin, The elements of statistical learning: data mining, inference and prediction, The Mathematical Intelligencer, 27 (2) (2005).
[17] C.M. Bishop, Pattern recognition and machine learning, springer, 2006.
[18] A. Wagdy, V. Garcia-Hansen, G. Isoardi, K. Pham, A parametric method for remapping and calibrating fisheye images for glare analysis, Buildings, 9 (10) (2019) 219.
[19] J.A. Jakubiec, K.V.D. Wymelenberg, M. Inanici, A. Mahic, Accurate measurement of daylit interior scenes using high dynamic range photography, in: Proceedings of the CIE 2016 Lighting Quality and Energy Efficiency Conference, 2016.
[20] J.A. Jakubiec, K. Van Den Wymelenberg, M. Inanici, A. Mahić, Improving the accuracy of measurements in daylit interior scenes using high dynamic range photography, in: Proceedings of Passive and Low Energy Architecture (PLEA) 2016 Conference, Los Angeles, CA, July 11–13, 2016.
[21] M. Inanici, Evaluation of high dynamic range photography as a luminance data acquisition system, Lighting Research and Technology, 38 (2) (2006) 123-134.
[22] M. Inanici, Evalution of High Dynamic Range Image-Based Sky Models in Lighting Simulation, LEUKOS, 7 (2) (2010) 69-84.
[23] A. Wagdy, V. Garcia-Hansen, G. Isoardi, A.C. Allan, Multi-region contrast method–A new framework for post-processing HDRI luminance information for visual discomfort analysis, in: PLEA 2017: Design to Thrive, Edinburgh, 2017.
[24] A. Wagdy, F. Fathy, A parametric approach for achieving optimum daylighting performance through solar screens in desert climates, Journal of Building Engineering, 3 (2015) 155-170.
[25] M. Curtin, Brute Force, Springer, Dordrecht, 2007.
[26] A. Wagdy, F. Fathy, A Parametric Approach for Achieving Daylighting Adequacy and Energy Efficiency by Using Solar Screens, in: PLEA 2016 - 36th International Conference on Passive and Low Energy Architecture, Los Angeles, California, USA, 2016.
[27] J. Wienold, C. REETZ, T. KUHN, Evalglare: a new RADIANCE-based tool to evaluate glare in office spaces, in: 3rd International Radiance Workshop, 2004.
[28] C. Pierson, M. Sarey Khanie, M. Bodart, J. Wienold, Discomfort Glare Cut-off Values from Field and Laboratory Studies, in: 29th Quadrennial Session of the CIE, Washington DC, USA, 2019, pp. 295-305.
[29] K. Van Den Wymelenberg, Patterns of occupant interaction with window blinds: A literature review, Energy and Buildings, 51 (2012) 165-176.
[30] J.D. Rodriguez, A. Perez, J.A. Lozano, Sensitivity Analysis of k-Fold Cross Validation in Prediction Error Estimation, IEEE Transactions on Pattern Analysis and Machine Intelligence, 32 (3) (2010) 569-575.
[31] D.W. Hosmer Jr, S. Lemeshow, R.X. Sturdivant, Applied logistic regression, John Wiley & Sons, 2013.
[32] B. Schölkopf, A.J. Smola, R.C. Williamson, P.L. Bartlett, New support vector algorithms, Neural computation, 12 (5) (2000) 1207-1245.
[33] C. Kurian, R. Aithal, J. Bhat, V. George, Robust control and optimisation of energy consumption in daylight—artificial light integrated schemes, 40 (1) (2008) 7-24.
[34] A. Mahić, K. Galicinao, K. Van Den Wymelenberg, A pilot daylighting field study: Testing the usefulness of laboratory-derived luminance-based metrics for building design and control, Building and Environment, (2017).
[35] C. Seiffert, T.M. Khoshgoftaar, J.V. Hulse, A. Napolitano, RUSBoost: Improving classification performance when training data is skewed, in: 2008 19th International Conference on Pattern Recognition, 2008, pp. 1-4.
[36] M. Elhenawy, M. Masoud, S. Glaser, A. Rakotonirainy, A new Approach to Improve the Topological Stability in Non-linear Dimensionality Reduction, IEEE Access, (2020) 1-1.